\documentclass[conference]{IEEEtran}

\usepackage{amssymb,amsmath}
\usepackage{cite}
\usepackage{graphicx}
\usepackage{psfrag}
\usepackage{url}
\usepackage[latin1]{inputenc}
\usepackage[absolute,overlay]{textpos}
\usepackage{gensymb}
\usepackage{cases}
\usepackage{graphicx}
\usepackage[font=footnotesize]{caption}
\usepackage[font=footnotesize]{subcaption}
\usepackage{float}

\usepackage{pgf}

\usepackage[nocomma]{optidef}

\usepackage{algorithm}
\usepackage{algorithmic}

\usepackage{amsthm}

\usepackage{booktabs}
\usepackage{color,soul}

\usepackage[bottom=0.73in,top=0.7in,left=0.57in,right=0.57in]{geometry} \textwidth=7.25in

\begin{document}

\title{A Learning-Based Trajectory Planning of Multiple UAVs for AoI Minimization in IoT Networks}
\author{
	\IEEEauthorblockN{Eslam Eldeeb, Dian Echevarr\'ia P\'erez, Jean Michel de Souza Sant'Ana, Mohammad Shehab, Nurul Huda\\
	Mahmood, Hirley Alves and Matti Latva-aho \\
	}
	\IEEEauthorblockA{Centre for Wireless 	Communications (CWC), University of Oulu, Finland \\
	Email: firstname.lastname@oulu.fi}
}
\maketitle

\vspace{0mm}
\begin{abstract}
Many emerging Internet of Things (IoT) applications rely on information collected by sensor nodes where the freshness of information is an important criterion. \textit{Age of Information} (AoI) is a metric that quantifies information timeliness, i.e., the freshness of the received information or status update. This work considers a setup of deployed sensors in an IoT network, where multiple unmanned aerial vehicles (UAVs) serve as mobile relay nodes between the sensors and the base station. We formulate an optimization problem to jointly plan the UAVs' trajectory, while minimizing the AoI of the received messages. This ensures that the received information at the base station is as fresh as possible. The complex optimization problem is efficiently solved using a deep reinforcement learning (DRL) algorithm. In particular, we propose a deep Q-network, which works as a function approximation to estimate the state-action value function. The proposed scheme is quick to converge and results in a lower AoI than the random walk scheme. Our proposed algorithm reduces the average age by approximately $25\%$ and requires down to $50\%$ less energy when compared to the baseline scheme.
\end{abstract}
\section{Introduction}

The Internet of Things (IoT) has enabled the deployment of sensor nodes (SN) to collect real-time information in various scenarios. This allows implementing many time-sensitive applications that operate on fresh information to improve the performance and the efficiency of the underlying use case. Use cases include, e.g., human safety applications such as intelligent 
transportation and prediction for physical safety, or event monitoring applications such as temperature and humidity sensing in smart agriculture~\cite{abd2019role}.

Meanwhile, the age of information (AoI) is a metric that quantifies the freshness of information about a remote system. 
AoI is defined as the time elapsed since the generation of the packet that was most recently delivered to the destination \cite{kadota2018scheduling}. Minimizing the AoI in IoT applications has attracted lots of attention recently in many applications. For instance, reducing the AoI in vehicular communications (\textit{e.g}., state of traffic lights, vehicles, and road sensor states, etc ...) could prevent the occurrence of accidents. There is also a special interest in scenarios with power-limited SNs, and their communication with the base station (BS) is difficult or even infeasible most times~\cite{GCZ+19_AoI_CR, HLC21_AoIminimization}. 
The SNs in such scenarios might not transmit the signals with sufficient power and hence will not achieve the signal-to-interference plus noise ratios (SINR) required to decode the data at the BS. 

A potential solution to solve the aforementioned problem is the use of mobile nodes, such as unmanned aerial vehicles (UAV), with low operational costs and flexible deployment capabilities to collect information from end devices and re-transmit it to the BS~\cite{HXQ+21_AoItrajectory}. In addition, UAVs also offer the possibility of reaching high altitudes while flying, hence, increasing the probability of line-of-sight (LOS) with both BS and ground users. Their hovering capabilities allow them to stay stationary at certain positions over the surface for a given time period \cite{mozaffari2019tutorial}. These characteristics make UAVs a suitable option for implementing of efficient AoI minimization methods. 

Determining the optimal UAVs trajectory to minimize the AoI considering the activation pattern of the IoT nodes is among the key design challenges. This has been addressed in the literature by formulating an optimization problems, which are then solved using optimal or heuristic algorithms, such as dynamic programming and ant colony optimization~\cite{HXQ+21_AoItrajectory}. 
However, the complexity of the original problem requires significant simplifications to switch it to a feasible optimization problem, reducing its practical appeal. An alternative option to find the optimal trajectory is the implementation of learning methods such as reinforcement learning (RL), which allows the agents (UAVs) to learn the environment and determine the optimal flight policy.

\subsection{Related Literature}
Several works have considered UAV-assisted communication for AoI minimization using learning methods~\cite{zhou2019deep, deep_us, deep_china, tong2020deep}. For instance, the work in ~\cite{zhou2019deep} investigated an online AoI-based trajectory planning for a UAV-assisted IoT network with unknown traffic generation and device topology. The AoI minimization problem was formulated as a Markov decision process (MDP) and solved using deep reinforcement learning (DRL). 
The authors in \cite{deep_us} considered discrete battery levels where one UAV assists the communication of sensors with energy constraints. They proposed an age-optimal policy for optimizing the UAV's flight trajectory and Sensors scheduling, and implemented it using DRL. The proposed algorithm showed better results in terms of average sum-AoI when compared to random walk (RW) and distance-based policies. A similar approach was presented in~\cite{deep_china} where the authors formulated an MDP to find the optimal trajectory of the UAV together with the optimal transmission scheduling of the SNs that minimizes the weighted AoI, with UAV battery guarantees. They proposed an UAV-assisted data collection algorithm that outperformed distance-based and AoI-based approaches. Similar to previous works, the authors of \cite{tong2020deep} addressed the minimization of the average AoI of all deployed sensors by optimizing the UAV's flight trajectory, while keeping the packet drop rate as low as possible. The SNs were assumed to sample information with either fixed or random rates. 
Authors cast the AoI minimization problem as a finite-horizon MDP and propose a DRL algorithm to find the optimal solution. However, most of these works assumed a single UAV, constrained UAV mobility options, and a limited number of SNs in their system model. 
\subsection{Contributions}
The major contributions of this paper are:
\begin{itemize}
    \item We address the problem of the average AoI minimization of randomly deployed IoT devices with multiple UAVs assisting the communication, unlike the single UAV scenario in most existing literature.
    \item We propose a DRL algorithm for finding the optimal policy for the UAV's trajectories with five and nine degrees of freedom 
    \item We show that our proposed DRL approach improves average AoI and energy consumption when compared to baseline approaches such as RW.
    \item We numerically validate the proposed algorithm with up to 10 deployed SNs, which represents an improvement over previous works where the number of IoT nodes is usually fewer.
\end{itemize}
Applications of our model include but are not limited to the field of environmental monitoring, e.g., UAV-based forest fire surveillance \cite{yuan2017fire}, where one or more UAVs might be deployed to collect real-time information which is critical in disaster prevention. 
\subsection{Outline} Section~\ref{system} presents the system model. The proposed DRL algorithm is detailed and numerically evaluated in Sections~\ref{DQN} and~\ref{results}, respectively. Finally, Section~\ref{conclusions} concludes the paper and discusses potential future works. 





\section{System Layout and Problem Formulation}\label{system}
\subsection{System Layout}


We consider a large area with a set $\mathcal{D}=\{1,2,\cdots,D\}$ of $D$ low-power single-antenna IoT devices randomly deployed in the 2D plane to monitor different physical processes as in \cite{deep_us,deep_china}. The location of each device $d \in \mathcal{D}$ is given by $L_d=(x_d,y_d)$. A BS is located at the center of the area (i.e, at $(0,0)$). A set $\mathcal{U} =\{1,2,\cdots,U\}$ of $U$ rotary-wing UAVs is dispatched to collect information from all the deployed devices by flying over different spots within the service area. The main objective is to gather information from the devices in a way which reduces the weighted sum of AoI while reducing the energy consumption for each IoT device. Each UAV then relays the information from the IoT devices to the BS at the center of the map. A set of UAV charging depots $\mathcal{C}=\{1,2,\cdots,C\}$ is conveniently deployed at fixed positions around this area (for example, at the corners). The position of each charging depot $c \in \mathcal{C}$ is given by $b_c = (x_c,y_c)$.

\begin{figure}[t!]
	\centering
	\includegraphics[width=0.99\columnwidth]{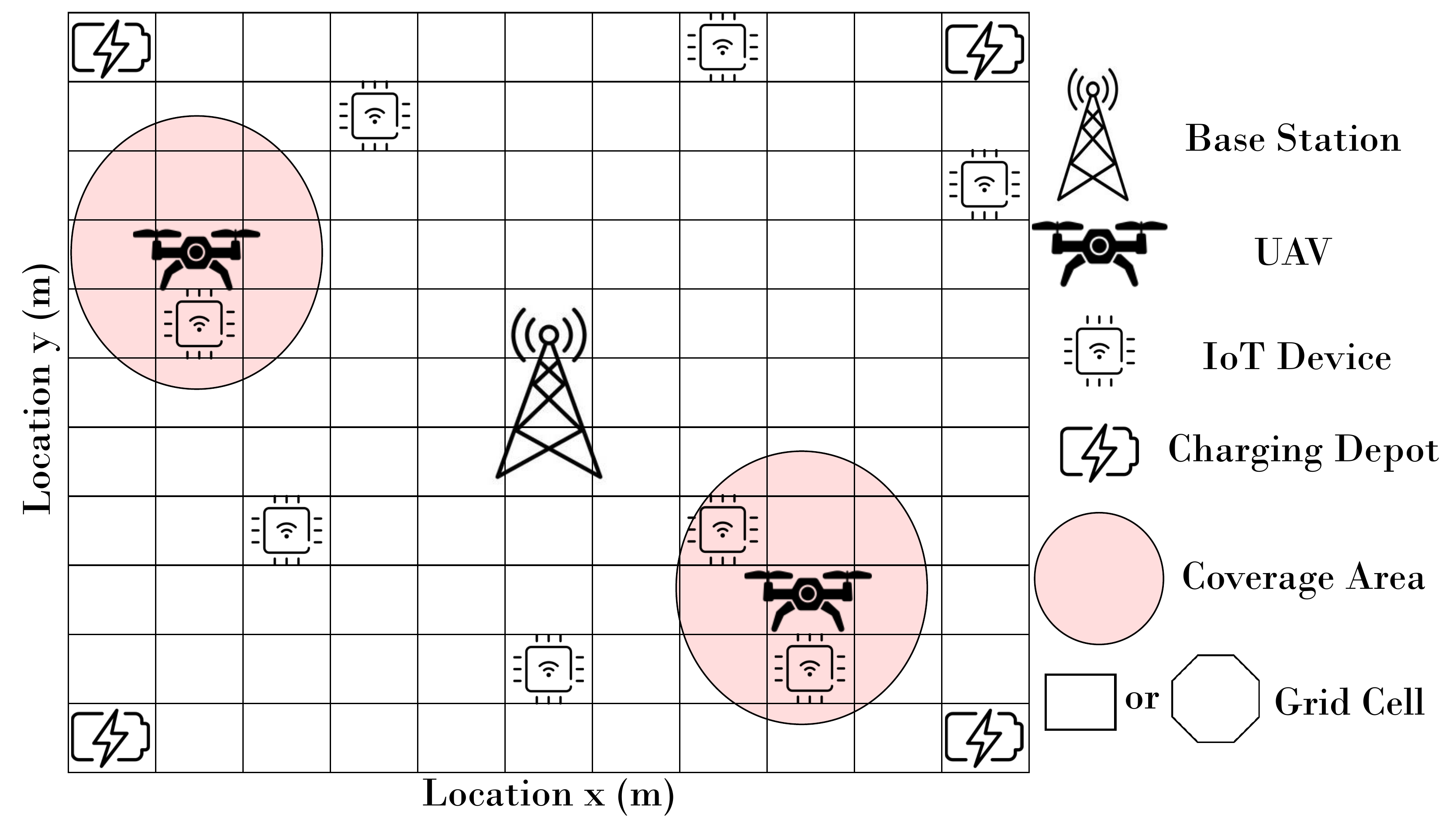}
	\caption{The system model comprises a set $\mathcal{D}$ of IoT devices served by a set $\mathcal{U}$ of rotary-wing UAV. Each UAV relays the information from the IoT devices to the BS in the middle of the map.} \vspace{-2mm}
	\label{Fig1}
\end{figure}

Without loss of generality, a discrete-time system is assumed, where time is divided into slots of unit length such that each time slot $t \geq 1$ corresponds to the time duration $[t - 1, t]$. The position of the UAV $u$ at time slot $t$ is fully given by its projection on the 2D plane $p_u(t)=(x_u(t),y_u(t))$ and its altitude $h_u$. It is reasonable to say that a UAV must start its trip from a charging depot and at the end of a trip, it should head for a charging depot to charge its battery before starting the next trip. Then, the movement of a UAV in one trip $u$ is described as a sequence of projections on the ground at each time slot $t$ such that $L_u(t) = [l_u(1), l_u(2), l_u(3),...,l_u(T_u)]$ with \{$l_u(1), l_u(T_u) \in \mathcal{C}\}$ denote the initial and final locations, respectively. For convenience, the area of interest is discretely divided into equally-sized small square/octagonal grids, where the positions of UAVs or IoT devices are considered to be constant anywhere inside one grid. The center of each grid is given by $(x_g,y_g)\in \mathcal{G}$, where $\mathcal{G}$ is the set containing the locations of the centers of each grid. The distance between the centers of two adjacent grids is $r_g$. Moreover, we set $\tau$ as the time required for the UAV to move from the center of one grid to another, which is defined as the ratio between $r_g$ and the UAV velocity. Let's also assume that the SNs follow a scheduling policy such that $w(t) \in \mathcal{W} = \{0,1,...,D\}$ where $w(t) = d$ means that node $d$ is scheduled to transmit at time slot $t$. The system model is illustrated in Fig. \ref{Fig1}.

The power consumption of the UAVs when moving or hovering is composed of three components modeled by \cite{zeng2019energy}
\begin{align}
P_u(V_t)=P_0& \left( 1+\frac{3V_t^2}{U_{tip}^2} \right)+P_1\left(\sqrt{1+\frac{V_t^4}{4\mu_0^4}}-\frac{V_t^2}{2\mu_0^2}\right)^\frac{1}{2} \nonumber\\
+&\frac{1}{2}d_0\rho s_0 B V_t^3,
\end{align}
where $P_0$ and $P_1$ represent the blade profile power and derived power when the UAVs are hovering, respectively. $V_t$ depicts the velocity of the UAVs and $U_{tip}$ the tip speed of the rotor blade, $\mu_0$ is the mean rotor induced velocity when hovering, $d_0$ represents the fuselage drag radio, $\rho$ is the air density. Meanwhile, $s_0$ represents the rotor solidity and $B$ the area of the rotor disk. Moreover, the UAVs consume energy when communicating with the BS. If we consider a battery capacity $E_{max,u}$ for the UAVs and discretize it in energy quanta $e_u$, then, the amount of energy per energy quantum is given by the ratio $E_{max,u}/e_{max,u}$. We denote the battery level of UAV $u$ at time slot $t$ as $e_u(t) \in \mathcal{E}_u = \{0,1,...,e_{u,max}\}$. Then, the energy consumption at a UAV required to relay an update packet is given by
\begin{align}
    e_u^R(t) = \frac{e_{max,u}}{E_{max,u}}E_u(t),
\end{align}
with \vspace{-2mm}
\begin{align}
    E_u(t) = \frac{\sigma^2}{g_{u,bs(t)}}\big(2^{\frac{M}{BW}}-1\big),
\end{align}
 where $g_{u,bs}(t)$ is given in \eqref{eq_3}, $M$ is the packet size of the sensor updates, $BW$ depicts the signal bandwidth and $\sigma^2$ the noise power. The energy consumption due to flying and hovering can also be expressed as 
 \begin{equation}
     e_u^F(V_t)=\frac{e_{max,u}}{E_{max,u}} P_u(V_t).
 \end{equation}
Since the energy consumed due to flying or hovering is considerably larger than the energy for packets relays, the number of energy quanta when discretizing the batteries must be set large enough in order not to overestimate the energy consumption due to communication with the BS. The battery evolution of the UAVs is given by 
\begin{equation}
	e_u{(t\!+\!1)}=
	\begin{cases}
		e_u(t)-\lceil e_u^F(V_t)+e_u^R(t) \rceil, & \text{if}   \ w(t) = d, \\
		e_u(t)-\lceil e_u^F(V_t)\rceil, & \text{otherwise}.  \end{cases}
\end{equation}
The ceiling operator guarantees a lower bound on the energy performance. The scheduling constraint $||p_u(t)-L_d||\leq R_d$ ensures that UAV $u$ is within the coverage radius $R_d$ of sensor $d$. $R_d$ is defined as
\begin{align}
    R_d = \Bigg(\cfrac{\beta_0P}{(2^{\frac{M}{BW}}-1)\sigma^2}-h_u^2\Bigg)^{1/2},
\end{align}
where $P$ represents the transmit power of a SN \cite{deep_china}.

We also assume the presence of LOS between the sensors and UAVs, and between the UAVs and BS, therefore, the channel gain between UAV $u$ and the BS at time slot $t$ is given by
\begin{align}
\label{eq_3}
    g_{u,bs}(t) = \beta_0d_{u,bs}^{-2} = \frac{\beta_0}{|h_u-h_{bs}|^2+||L_u(t)||^2},
\end{align}
where $\beta_0$ is the channel gain at the reference distance of 1 m and $h_{bs}$ represents the height of the antennas at the BS\cite{deep_us}. We use the AoI as a metric for measuring the freshness of information, which is defined as the time elapsed since the last update packet received at a UAV was generated. Particularly, we define AoI as 
\begin{equation}
	A_d(t+1) =
	\begin{cases}
		1, & \quad \text{if} \ w(t) = d, \\
		\text{min}\{A_{d,max}(t),A_{d}(t) + 1\}, & \quad \text{otherwise}, 
	\end{cases}
\end{equation}
which ensures that the AoI of SN $d$ is set to one if it generates an update packet at time slot $t$ or is increased by one otherwise. Here, $A_{d,max}$ denotes the maximum allowed AoI, which is relatively large.

\subsection{State and Action Spaces}
 The state space of the system at time slot $t$ is defined as $s(t) = (\boldsymbol{l}(t),\boldsymbol{A}(t), \boldsymbol{\Delta}(t))$ where:
\begin{itemize}
    \item $\boldsymbol{l}(t)$  is a vector containing the position  of each UAV $l_u(t)\in\mathcal{P}$ at time slot $t$.
    \item $\boldsymbol{A}(t) = (A_1(t), A_2(t),...,A_N(t))$ ($N<D$) contains the AoI of all SNs under the coverage of the UAVs, where $A_N(t)\in\mathcal{A} = [1,2,...,A_{N,max}]$.
    \item $\boldsymbol{\Delta}(t) = (\Delta_1(t), \Delta_2(t),...,\Delta_U(t))$ with $\Delta_U(t)\in \mathcal{G}$, is a vector that contains the difference between the battery status of each UAV  and both the required energy to arrive to the nearest charging depot $c \in \mathcal{C}$ and the energy consumed by packet relays considering the worst case when the UAVs relay packets in every time slot $t$.
\end{itemize}
Finally, the state space of the system is  given by $\mathcal{S}= \mathcal{P}^{U}\times\mathcal{A}^N\times\mathcal{G}^U$.

The action space at time slot $t$ is determined by the movement of the UAVs $v_u(t)$ and the scheduling policy, i.e, $a(t) = (v_u(t),w(t))$. Here, $v_u(t)\in \mathcal{V}$ has 9 directional actions as detailed in \eqref{eqn:directions}. Then, the action space is given by $\mathcal{B} = \mathcal{V}^U\times\mathcal{W}^U$. 

The movement direction of each UAV at a given time slot is modeled as
\begin{equation} \label{eqn:directions}
	l_u(t+1)=
	\begin{cases}
		l_u(t)+(0,r), & \quad v_u(t)=\text{North}, \\
		l_u(t)-(0,r), & \quad v_u(t)=\text{South}, \\
		l_u(t)+(r,0), & \quad v_u(t)=\text{East}, \\
		l_u(t)-(r,0), & \quad v_u(t)=\text{West}, \\
		l_u(t)+(\frac{r}{\sqrt{2}},\frac{r}{\sqrt{2}}), & \quad v_u(t)=\text{Northeast}, \\
		l_u(t)+(-\frac{r}{\sqrt{2}},\frac{r}{\sqrt{2}}), & \quad v_u(t)=\text{Northwest}, \\
		l_u(t)+(\frac{r}{\sqrt{2}},-\frac{r}{\sqrt{2}}), & \quad v_u(t)=\text{Southeast}, \\
		l_u(t)+(-\frac{r}{\sqrt{2}},-\frac{r}{\sqrt{2}}), & \quad v_u(t)=\text{Southwest}, \\
		l_u(t), & \quad \text{Hovering}. \\
	\end{cases}
\end{equation}

The reward system is defined to minimize the weighted-sum of the age of information for all devices. We define the immediate reward $r_u$ for the $u$ UAV at time instant $t$ as
\begin{equation}
r_u(t) = 
    \begin{cases}
    - \sum_{d = 1}^{D}\theta_d A_d(t) \: - \mathrm{z}, & \quad
    \text{if } d_{u,d} > R_d, \\
    - \sum_{d = 1}^{D}\theta_d A_d(t), & \quad 
    \text{otherwise}, \\
    \end{cases}
\end{equation}
where $\theta_d$ is a weight that denotes the importance of sensor $d$, $\mathrm{z}$ represents energy penalty from allocating a device outside the coverage radius of the UAV, and $d_{u,d}$ is the distance between the UAV and the chosen SN $d$. In addition, we define an episodic model, where each episode starts by having all UAVs at one of the charging depots. We assume a centralized learning that takes place at the BS, where all the states and actions are shared among all the UAVs. Therefore, an episode ends by having at least 1 UAV with an amount of energy quanta $e_u$ less than or equal the energy threshold $e_{th}$, where the UAV takes the shortest path to one of the recharging depots.

\subsection{Problem Formulation}\label{INF}
We aim at minimizing the weighted average AoI of all SNs in the network by jointly finding the optimal trajectories of all deployed UAVs. We can proceed to formulate the optimization problem as follows \vspace{-2mm}
\begin{subequations}\label{P1}
	\begin{alignat}{2}
	\mathbf{P1:}\qquad &\underset{\boldsymbol{l}(t)}{\min}       &\ \ \ & \frac{1}{T}\sum_{t=1}^T\sum_{d = 1}^{D}\theta_d A_d(t),\label{P1:a}
	\ \\
	&\text{s.t.}   &      & \sum_t^{T_u}P_u(V_t)\leq e_u(t), \label{P1:b}\\
		& & & l_u(1) = b_{c,u}, \label{P1:c}
	\end{alignat}
\end{subequations}
where $b_{c,u}$ are the coordinates of the charging depot where UAV $u$ is going to take off. \eqref{P1:a} represents the weighted average AoI of all nodes in the network, \eqref{P1:b} ensures that the UAVs will be able to reach a charging depot before running out of energy, and \eqref{P1:c} establishes the initial position where the UAVs take off from. It is worth noting that \eqref{P1:b} depends on $T_u$, which means that the period that the UAVs will be flying/hovering are different and they can arrive at the charging depots in different time slots, considering that they all take off at the same time. The optimization problem~\eqref{P1} is a non-linear integer programming optimization problem whose complexity grows with the number of deployed devices. In order to solve this problem in an efficient and feasible manner, we propose a DRL-based approach with the use of action/state spaces and the reward system as defined in the previous subsection.  
\section{DRL Approach}\label{DQN}
In regular RL problems, the goal of the agent is to find the best policy to follow while being at each state~\cite{ADB+17_DRLsurvey}. The state-action value function $Q_\pi(s,a)$ is one of the key functions in RL problems to find the optimal policy. It represents the expected reward of taking an action $a$ at state $s$ then following a policy $\pi$. The state-action value function at time instant $t$ is updated as follows 
\begin{align}
& Q\left(s\left(t\right),a\left(t\right)\right) = \:  Q\left(s\left(t\right),a\left(t\right)\right) + \nonumber\\
&\alpha \:  \left(r\left(t\right) +  \gamma \: \max_a Q\left(s\left(t+1\right),a\right)
-Q\left(s\left(t\right),a\left(t\right)\right)\right),
\end{align} 
where $\alpha$ is the learning rate, $r(t)$ is the immediate reward, $\gamma \: Q\left(s\left(t+1\right),a\left(t+1\right)\right)$ is the discounted state-action value at time instant $t+1$, and $\gamma$ is the discount factor. In addition, an exploration rate $\epsilon$ is defined, where the agent selects a random action with probability $\epsilon$ and selects the greedy action (the one that maximizes the state-action value) with probability $1-\epsilon$. The value of $\epsilon$ decays as the learning progresses. Hence, random actions are more likely to be chosen at the beginning of training to explore the state space, whereas it is better to follow the best policy after relatively long training period. In our model, the UAV experiences a large dimension state space, which is almost a continuous state space. To overcome the dimensionality issue, we propose a deep Q-network (DQN), which works as a function approximation to estimate the action-value function (Q-function).

In DQNs, there are two implemented neural networks \cite{DQNs}. The first network estimates the Q-function, whereas the second network is called the target network, which estimates the target Q-function. In the context of DQNs, there are two major strategies that improve the learning rate: fixed Q-targets and experience replay. The DQN updates the Q-function estimator network, while keeping the weights of the target network fixed, where they are only updated every specified number of steps to utilize the fixed Q-targets. In addition, the DQN stores the experience $(s\left(t+1\right),a\left(t+1\right),r\left(t+1\right),s\left(t+1\right))$ in a buffer memory, where a small batch from this buffer is sampled randomly to train the neural network. The experience replay utilizes the past experience and also breaks the correlation behavior of the samples, where the states $s_t$ and $s_{t+1}$ are highly correlated. Applying fixed Q-targets and experience replay strategies speed up the learning process and guarantee finding the optimal policy. 

\section{Numerical Analysis}\label{results}
\begin{figure*}
\centering
\begin{subfigure}{\columnwidth}
  \centering
  \includegraphics[]{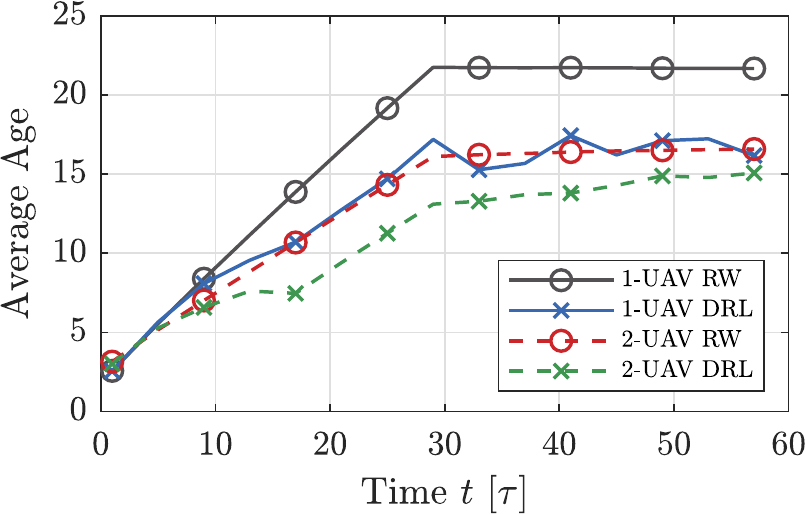}
  \caption{5-Directions Model.}
  \label{Fig2}
\end{subfigure}%
\begin{subfigure}{\columnwidth}
  \centering
  \includegraphics[]{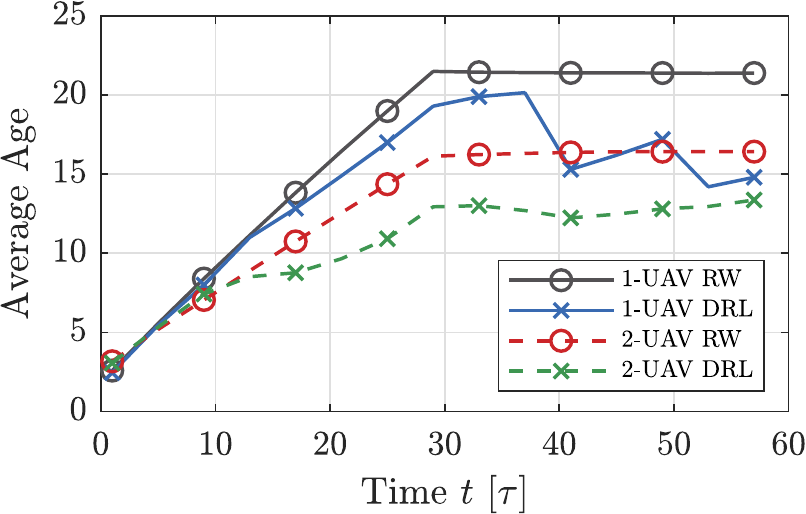}
  \caption{9-Directions Model.}
  \label{Fig3} 
\end{subfigure}
\caption{Average age of information for the RW and the proposed DRL approach with 1 and 2 UAVs serving 5 IoT devices using different direction models. Time is measured in multiples of $\tau$.}
\label{fig:ranges}
\end{figure*}

In this section, we present the numerical results of our proposed DRL scheme and compare them to a baseline RW model. First, consider a grid world of $1100$ m $\times$ $1100$ m, which is divided into $11 \times 11$ grids. The simulation parameters regarding communication for the UAV model are in Table \ref{tab:uav}. We use Pytorch to implement the proposed DQN, which comprises an input layer, which has the size of the state space, two fully connected hidden layers in case of one UAV and five fully connected hidden layers in case of two UAVs, and the output layer with the size of the action space. Table \ref{tab:parameters} summarizes the DQN setup settings. In addition, we use a single NVIDIA Tesla V100 GPU and 10 GB of RAM 
The one UAV case requires around $10000$ episodes (40 minutes) to converge to the minimum training loss, which corresponds to the optimal policy, whereas the two UAVs case requires around $25000$ episodes (100 minutes). We present two models to evaluate the trained UAVs performances. The first model (5-directions model) allows the UAV to move in five directions (North, South, East, West and Hovering), whereas the second model, as mentioned in \eqref{eqn:directions}.
\begin{table}[t!]
\centering
\caption{UAV model parameters}
\label{tab:uav}
\begin{tabular}{@{}cc@{}}
\toprule
\textbf{Parameter}                                    & \textbf{Value} \\ \midrule
Battery capacity $E_{max,u}$                          & 10000          \\
Energy per quantum $e_{max,u}$                        & 200            \\
Maximum allowed AoI $A_{d,max}$                       & 30             \\
Channel gain at 1 m $\beta_0$                         & 30 dB          \\
Height of the UAV $h_u$
                        & 100 m \\
distance between cells centers $r_g$                         & 100 m          \\
Bandwidth $BW$                                        & 1 MHz          \\
Packet size $M$                                       & 5 Mb           \\
Noise power $\sigma^2$                                & -100 dBm       \\
Number of charging depots $\mathcal{C}$               & 4              \\
UAV speed $V_t$                                       & 25 m/s         \\
Tip speed $U_{tip}$                                   & 120 m/s        \\
Air density $\rho$                                    & 1.225 kg/m$^3$ \\
Blade power profile $P_0$                             & 99.66 W        \\
Hovering power profile $P_1$                          & 120.16 W       \\
Fuselage drag ration $d_0$                            & 0.48           \\
Rotor solidity $s_0$                                  & 0.0001         \\
Area of rotor disk $B$                                & 0.5 s$^2$       \\
Mean rotor induced velocity in hover $\mu_0$          & 0.002 m/s      \\
SN transmit power $P$                                 & 0.003 W        \\
Base station height $h_{bs}$                          & 15 m           \\
Energy penalty $\mathrm{z}$ & 5 \\
\bottomrule
\end{tabular}
\end{table}

\begin{table}[!h]
\centering
\caption{DQN parameters used on simulations for 1 and 2 UAVs.}
\label{tab:parameters}
\begin{tabular}{@{}ccc@{}}
\toprule
\textbf{Parameter}                                                    & \textbf{1 UAV}                                                                                         & \textbf{2 UAVs}                                                                                         \\ \midrule

\begin{tabular}[c]{@{}c@{}}Neural network\\ architecture\end{tabular} & \begin{tabular}[c]{@{}c@{}}2 hidden layers\\ (64,64)\end{tabular}                                      & \begin{tabular}[c]{@{}c@{}}5 hidden layers\\ (64,128,256,128,128)\end{tabular}                         \\
Episodes                                                    & 50000                                                                                                  & 100000                                                                                                  \\
Batches                                                               & 64                                                                                                     & 128                                                                                                    \\
Learning rate $\alpha$                                                         & 0.0004                                                                                                 & 0.0004                                                                                                 \\
Initial $\epsilon$                                                         & 1                                                                                                & 1                                                                                                \\
$\epsilon$-decay                                                         & 0.995                                                                                                 & 0.995                                                                                                 \\
Discount factor $\gamma$                                                       & 0.99                                                                                                   & 0.99                                                                                                   \\
Replay buffer size                                                           & 100000                                                                                                 & 1000000                                                                                                \\
Optimizer                                                          & Adam                                                                                                 & Adam                                            \\
Activation function                                                          & ReLU                                                                                                 & ReLU                                            \\

 \bottomrule
\end{tabular}
\end{table}

Fig.~\ref{Fig2} presents the average age for RW and the proposed DRL approach with one and two UAVs using the 5-directions model, whereas Fig.~\ref{Fig3} presents the 9-directions model. We consider 5 IoT devices in both cases. Both figures show that the proposed DRL approach results in a better performance, i.e., reduced average AoI over the baseline random walk. When comparing the average AoI of the DRL scheme with one and two UAVs, we notice that there is a considerable gap at the beginning, which reduces with time. This is because, with the considered number of few IoT devices in the small grid world, the proposed DRL approach can learn the age-optimal scheduling policy efficiently, even for the case of a single UAV (as indicated in Fig.~\ref{Fig3}, where the corresponding age decreases after a certain point). The same does not occur with the random walk, where it seems always necessary to add more UAVs to increase the performance. Moreover, the average age seems to increase linearly in the beginning of the episodes as the individual ages are initialized as 1, where the devices experience low ages and their ages start to grow. After around $30$ time instants, the average age seems to converge to the true average age and becomes steady until the end of the episode. In addition, we can observe that using the optimal policy for the 9-directions model reduces the age experienced by the devices over the 5-directions model. With the 9-directions model, the UAV has more flexibility to reach the optimized destination faster than the 5-directions model.

\begin{figure}[t!]
	\centering
	\includegraphics[]{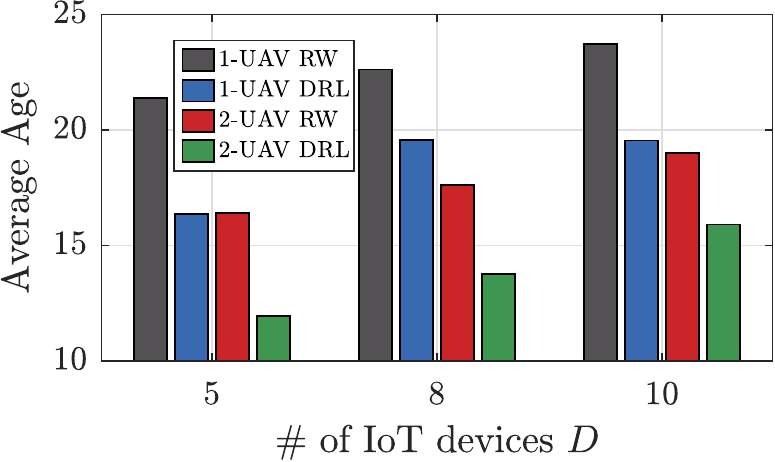}
	\caption{Average age of information for the RW and the proposed DRL approach with 1 and 2 UAVs using the 9-directions model as a function of the number of IoT devices $D$. 
	}.  \vspace{-2mm}
	\label{Fig5}
\end{figure}

Fig.~\ref{Fig5} shows the average AoI (i.e, final average age of the simulation time) regarding the number of IoT devices. We can see that the age increases with the number of IoT devices and the proposed DRL scheme outperforms the RW policy. In addition, the reduction in the age is quite significant while using DRL scheme over the RW policy when the number of IoT devices increases. The 1-UAV DRL almost achieves the same average AoI of the 2-UAV RW policy at $D = 10$. Since the UAV serves only 1 device at each time instant, the devices have to wait longer period until being served by the UAV, which increases the age in case of a large number of IoT devices. Therefore, the larger the deployment, the more significant the reduction of the age of the DRL policy compared to the baseline RW policy.

\begin{figure}[t!]
	\centering
	\includegraphics[]{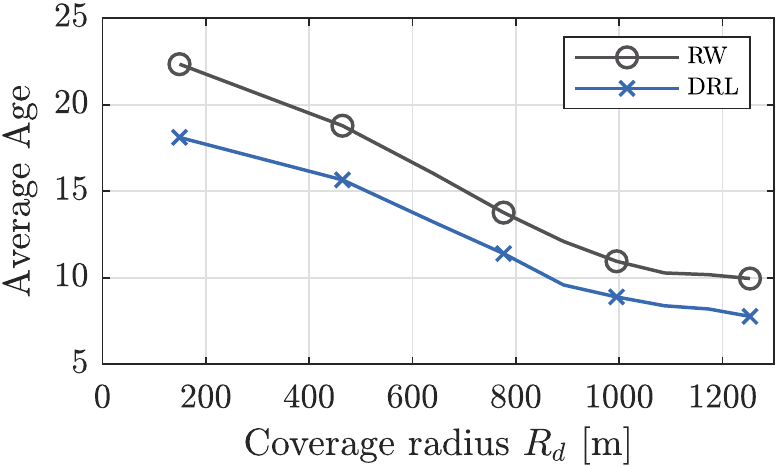}
	\caption{Total average age of information as a function of coverage radius for RW and the proposed DRL approach with 1 UAV serving 8 IoT devices using the 5-directions model.}
	\vspace{-2mm}
	\label{Fig6}
\end{figure}

In Fig.~\ref{Fig6}, 2 UAVs are serving 8 SNs using the 5-directions model while adjusting the allowed transmission power for the proposed DRL scheme. It shows the effect of the transmission power on the average age of information. As the transmission power increases, the coverage radius increases and it becomes easier for the UAV to receive information from any device in the grid world without energy penalty. Therefore, note that the age decreases as the coverage radius increases. Notice that the proposed DRL scheme outperforms the RW scheme even for high transmission power and the coverage radius of the UAV.

\begin{figure}[t!]
	\centering
	\includegraphics[]{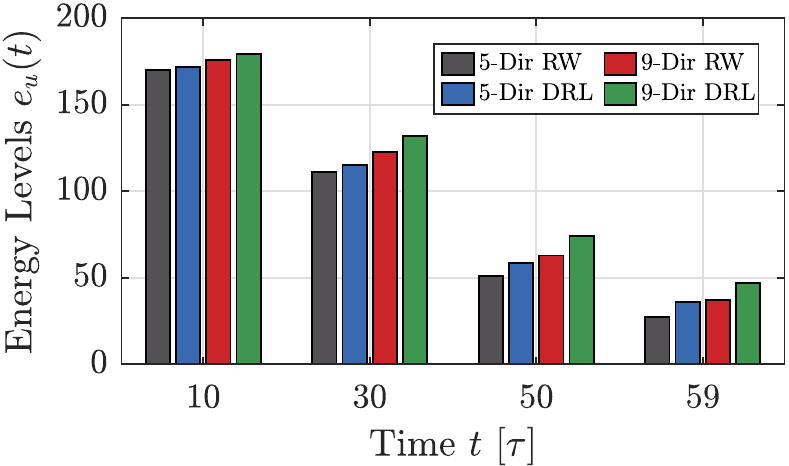}
	\caption{Available number of energy levels at the UAV for the random walk and the proposed DRL approach with 2 UAVs serving 5 IoT devices using the 5-directions model and the 9-directions model.} \vspace{-2mm} 
	\label{Fig7}
\end{figure}

Finally, Fig.~\ref{Fig7} illustrates the energy consumption for the proposed DRL scheme compared to the RW scheme for 2 UAVs serving 5 IoT devices using both the 5-directions model and the 9-directions model. The available energy levels presented is the average energy levels of the two UAVs. The DRL scheme finds the optimal policy that can achieve the best combination between the desired low age and saving UAV energy before recharging. As we noticed in \ref{Fig2} and \ref{Fig3}, the 9-directions model outperforms the 5-directions model in terms of average AoI. In addition, the 9-directions model also overpasses the 5-directions model by saving more energy levels as the 9-directions model has more flexibility in movement and can save time and energy by reaching the optimized location faster than the 5-directions model. It can be observed that the available energy levels $e_u(t)$ of the DRL scheme using the 9-directions model is almost double the available energy levels of the RW scheme using the 5-directions model after 59 time instants.


\section{Conclusions}\label{conclusions}
In this paper, we consider a network of deployed sensors in an IoT network where multiple UAVs serve as mobile relay nodes with LOS connectivity between the sensors and the UAVs. We formulate an optimization problem to plan the trajectory of the UAVs, while minimizing the AoI of the received messages and considering the energy consumption of the UAV nodes. We address the problem by proposing a DRL algorithm for finding the optimal policy for the UAV's trajectories with nine movement directions of the UAVs at each time instant. Our proposed approach provides better results in terms of average AoI and energy consumption when compared to the RW policy as a baseline scheme. In particular, our proposed algorithm reduces the average AoI by more than $25\%$ for the 9-direction model with one UAV and $10$ SNs. In addition, it also requires significantly less energy compared to the baseline scheme under all scenarios. This work would pave the road towards the deployment of UAV swarms to serve massive IoT scenarios. Thus, we plan to extend this work by solving the problem of high dimensionality in the action space, thus applying it to a massive IoT deployment scenario where we could train a UAV swarm to serve many IoT devices.

\section*{Acknowledgments} \vspace{1mm}
This work is partially supported by Academy of Finland, 6G Flagship program (Grant no. 346208) and FIREMAN (Grant no. 326301), and the European Commission through the Horizon Europe project Hexa-X (Grant Agreement no. 101015956).


%
%

%
\bibliographystyle{IEEEtran}
\bibliography{di}
\end{document}